%% file: acl_latex.tex
\pdfoutput=1

\documentclass[11pt]{article}

\usepackage[]{acl2023}

\usepackage{xcolor, soul}
\PassOptionsToPackage{dvipsnames}{xcolor}
\usepackage{times}
\usepackage{latexsym}
\usepackage{caption}
\usepackage{changes}
\usepackage{subcaption}
 \usepackage{colortbl} 
\usepackage[T1]{fontenc}
\usepackage{nicefrac}
\usepackage[utf8]{inputenc}

\usepackage{microtype}

\definecolor{HallRedShade}{HTML}{fdf1ec}
\definecolor{HallRedText}{HTML}{b76039}
\sethlcolor{HallRedShade}

\definecolor{citeblue}{HTML}{120d77}

\usepackage{titlesec}
\titlespacing{\paragraph}{%
  0pt}{
  0.45\baselineskip}{
  0.5em}

\title{Optimal Transport for Unsupervised Hallucination Detection \\ in Neural Machine Translation} 

\usepackage[misc]{ifsym}
\author{Nuno M. Guerreiro$^{1,2}$ \quad Pierre Colombo$^{4}$ \quad Pablo Piantanida$^{5}$ \quad André F. T. Martins$^{1,2,3}$ \\
$^1$Instituto de Telecomunicações, Lisbon, Portugal \\
$^2$Instituto Superior T\'ecnico \& LUMLIS (Lisbon ELLIS Unit), University of Lisbon, Portugal \\ $^3$Unbabel, Lisbon, Portugal\\
$^4$MICS, CentraleSupélec, Université Paris-Saclay\\
$^5$ILLS - CNRS, CentraleSupélec\\
\small \url{miguelguerreironuno@gmail.com}}

\input{math_commands.tex}

\usepackage{microtype}
\usepackage{inconsolata}
\usepackage{amsmath}
\usepackage{amssymb}
\usepackage{amsthm}
\usepackage{bbm}

\renewcommand\sfdefault{cmss}
\usepackage[nocomma]{optidef}

\usepackage{makecell}
\usepackage{multicol}
\usepackage{float}
\usepackage{lscape}
\usepackage{multirow}
\usepackage{tabularx}
\usepackage{booktabs}
\usepackage{colortbl}
\usepackage{enumitem}
\usepackage{algorithm}
\usepackage{algorithmic}
\usepackage{longtable}
\usepackage{tabu}
\usepackage{longtable}
\usepackage{arydshln}

\makeatletter
\def\adl@drawiv#1#2#3{%
        \hskip.5\tabcolsep
        \xleaders#3{#2.5\@tempdimb #1{1}#2.5\@tempdimb}%
                #2\z@ plus1fil minus1fil\relax
        \hskip.5\tabcolsep}
\newcommand{\cdashlinelr}[1]{%
  \noalign{\vskip 1.3pt
           \global\let\@dashdrawstore\adl@draw
           \global\let\adl@draw\adl@drawiv}
  \cdashline{#1}[.4pt/2pt]
  \noalign{\global\let\adl@draw\@dashdrawstore
           \vskip 1.3pt}}
\makeatother

\begin{document}
\maketitle
\begin{abstract}
Neural machine translation (NMT) has become the de-facto standard in real-world machine translation applications. However, NMT models can unpredictably produce severely pathological translations, known as hallucinations, that seriously undermine user trust. It becomes thus crucial to implement effective preventive strategies to guarantee their proper functioning. In this paper, we address the problem of hallucination detection in NMT by following a simple intuition: as hallucinations are detached from the source content, they exhibit cross-attention patterns that are statistically different from those of good quality translations. We frame this problem with an optimal transport formulation and propose a fully unsupervised, plug-in detector that can be used with any attention-based NMT model. Experimental results show that our detector not only outperforms all previous model-based detectors, but is also competitive with detectors that employ external models trained on millions of samples for related tasks such as quality estimation and cross-lingual sentence similarity.
\end{abstract}

\section{Introduction}
Neural machine translation (NMT)  has achieved tremendous success \citep{transformer_vaswani, kocmi-EtAl:2022:WMT}, becoming the mainstream method in real-world applications and production systems for automatic translation. Although these models are becoming evermore accurate, especially in high-resource settings, they may unpredictably produce \textit{hallucinations}. These are severely pathological translations that are detached from the source sequence content
\citep{lee2018hallucinations, muller-etal-2020-domain, raunak-etal-2021-curious, guerreiro-etal-2023-looking}. Crucially, these errors have the potential to seriously harm user trust in hard-to-predict ways~\cite{perez2022red}, hence the evergrowing need to develop security mechanisms. One appealing strategy to address this issue is to develop effective on-the-fly detection systems.

In this work, we focus on leveraging the cross-attention mechanism to develop a novel hallucination detector. This mechanism is responsible for selecting and combining the information contained in the source sequence that is relevant to retain during translation. Therefore, as hallucinations are translations whose content is detached from the source sequence, it is no surprise that connections between \textit{anomalous} attention patterns and hallucinations have been drawn before in the literature \citep{berard-etal-2019-naver, raunak-etal-2021-curious, altijavier2022}. These patterns usually exhibit scattered source attention mass across the different tokens in the translation (e.g. most source attention mass is concentrated on a few irrelevant tokens such as punctuation and the end-of-sequence token). Inspired by such observations, previous work has designed \textit{ad-hoc} heuristics to detect hallucinations that specifically target the anomalous maps. While such heuristics can be used to detect hallucinations to a satisfactory extent~\citep{guerreiro-etal-2023-looking}, we argue that a more theoretically-founded way of using anomalous attention information for hallucination detection is lacking in the literature.

Rather than aiming to find particular patterns, we go back to the main definition of hallucinations and draw the following hypothesis: as hallucinations---contrary to good translations---are not supported by the source content, they may exhibit cross-attention patterns that are statistically different from those found in good quality translations. Based on this hypothesis, we approach the problem of hallucination detection as a problem of anomaly detection with an \textbf{optimal transport (OT) formulation}~\citep{kantorovich2006translocation, peyre2019computational}. Namely, we aim to find translations with source attention mass distributions that are highly distant from those of good translations. Intuitively, the more distant a translation's attention patterns are from those of good translations, the more \textbf{anomalous} it is in light of that distribution. 

Our key contributions are:
\begin{itemize}
\item We propose an OT-inspired fully unsupervised hallucination detector that can be plugged into any attention-based NMT model;
\item We find that the idea that attention maps for hallucinations are anomalous in light of a reference data distribution makes for an effective hallucination detector;
\item We show that our detector not only outperforms all previous model-based detectors, but is also competitive with external detectors that employ auxiliary models that have been trained on millions of samples.\footnote{Our code and data to replicate our experiments are available in \url{https://github.com/deep-spin/ot-hallucination-detection}.}
\end{itemize}

\section{Background}
\subsection{Cross-attention in NMT models}
A NMT model $\mathcal{M}$ defines a probability distribution $p_{\bm{\theta}}(\bm{y}|\bm{x})$ over an output space of hypotheses $\mathcal{Y}$ conditioned on a source sequence $\bm{x}$ contained in an input space $\mathcal{X}$. In this work, we focus on models parameterized by an encoder-decoder transformer model~\citep{transformer_vaswani} with a set of learned weights $\bm{\theta}$. In particular, we will look closely at the cross-attention mechanism, a core component of NMT models that has been extensively analysed in the literature~\citep{Bahdanau2014NeuralMT, raganato-tiedemann-2018-analysis, kobayashi-etal-2020-attention, ferrando-costa-jussa-2021-attention-weights}. This mechanism is responsible for computing, at each generation step, a distribution over all source sentence words that informs the decoder on the relevance of each of those words to the current translation generation step. We follow previous work that has drawn connections between hallucinations and cross-attention~\citep{berard-etal-2019-naver, raunak-etal-2021-curious}, and focus specifically on the last layer of the decoder module. Concretely, for a source sequence of arbitrary length $n$ and a target sequence of arbitrary length $m$, we will designate as $\bm{\Omega} \in [0,1]^{m\times n}$ the matrix of attention weights that is obtained by averaging across all the cross-attention heads of the last layer of the decoder module. Further, given the model $\mathcal{M}$ we will designate~$\bm{\pi}_{\mathcal{M}}(\bm{x}) := \frac{1}{m} \left[\bm{\Omega}(\bm{x})\right]^\top \bm{1} \in  \triangle_n$ as the source (attention) mass distribution computed by $\mathcal{M}$ when $\bm{x}$ is presented as input, where $\triangle_n = \{\bm{p} \in \mathbb{R}^n \,|\, \bm{p} \ge \bm{0}, \, \bm{1}^\top \bm{p} = 1\}$ is the $(n-1)$-dimensional probability simplex.

\subsection{Optimal Transport Problem and Wasserstein Distance}\label{subsec:wasserstein}
The first-order Wasserstein distance between two arbitrary probability distributions $\bm{\mu} \in \triangle_n$ and $\bm{\nu} \in \triangle_m$ is defined as 
\begin{equation}
    W(\bm{\mu}, \bm{\nu}) = \inf_{\bm{\gamma} \in \Pi(\bm{\mu}, \bm{\nu})} \mathbb{E}_{(u,v) \sim \bm{\gamma}}\, \left[c(u,v)\right],
    \label{eq:wass_dist}
\end{equation}
where $c: [n] \times [m] \rightarrow \mathbb{R}_0^+$ is a cost function,\footnote{We denote the set of indices $\{1, \ldots, n\}$ by $[n]$.} and $\Pi(\bm{\mu}, \bm{\nu}) = \{ \bm{\gamma} \in \triangle_{n \times m} : \bm{\gamma} \bm{1} = \bm{\mu}; \bm{\gamma}^\top \bm{1} = \bm{\nu}\}$\footnote{We extend the simplex notation for matrices representing joint distributions, $\triangle_{n \times m} = \{\bm{P} \in \mathbb{R}^{n \times m} : \bm{P} \ge~\bm{0}, \bm{1}^\top \bm{P} \bm{1} = 1\}$.} is the set of all joint probability distributions whose marginals are $\bm{\mu}$, $\bm{\nu}$. The Wasserstein distance arises from the method of optimal transport (OT) \citep{kantorovich2006translocation, peyre2019computational}: OT measures distances between distributions in a way that depends on the geometry of the sample space. Intuitively, this distance indicates how much probability mass must be transferred from $\bm{\mu}$ to $\bm{\nu}$ in order to transform $\bm{\mu}$ into $\bm{\nu}$ while minimizing the transportation cost defined by $c$.

A notable example is the Wasserstein-1 distance, $W_1$, also known as Earth Mover's Distance (EMD), obtained for $c(u,v) = \|u-v\|_1$. The name follows from the simple intuition: if the distributions are interpreted as ``two piles of mass'' that can be moved around, the EMD represents the minimum amount of ``work'' required to transform one pile into the other, where the work is defined as the amount of mass moved multiplied by the distance it is moved.

Although OT has been explored for robustness~\cite{paty2019subspace, staerman2021ot} and out-of-distribution detection~\cite{wang2021wood,yan20212d,cheng2022hyperspectral} in computer vision, the use of OT for anomaly detection in NLP applications remains largely overlooked.

\subsection{The problem of hallucinations in NMT}
\label{ssec:hallucinationsinnmt}
Hallucinations are translations that lie at the extreme end of NMT pathologies~\citep{raunak-etal-2021-curious}. Despite being a well-known issue, research on the phenomenon is hindered by the fact that these translations are rare, especially in high-resource settings. As a result, data with hallucinations is scarce. To overcome this obstacle, many previous studies have focused on amplified settings where hallucinations are more likely to occur or are easier to detect. These include settings where (i)~perturbations are induced either in the source sentence or in the target prefix~\citep{lee2018hallucinations, muller-sennrich-2021-understanding, voita-etal-2021-analyzing, altijavier2022}; (ii)~the training data is corrupted with noise~\citep{raunak-etal-2021-curious}; (iii)~the model is tested under domain shift~\citep{wang-sennrich-2020-exposure, muller-etal-2020-domain}; (iv)~the detectors are validated on artificial hallucinations~\citep{zhou-etal-2021-detecting}. Nevertheless, these works have provided important insights towards better understanding of the phenomenon. For instance, it has been found that samples memorized by an NMT model are likely to generate hallucinations when perturbed~\citep{raunak-etal-2021-curious}, and hallucinations are related to lower source contributions and over-reliance on the target prefix~\citep{voita-etal-2021-analyzing, altijavier2022}. 

In this work, we depart from artificial settings, and focus on studying hallucinations that are \textit{naturally} produced by the NMT model. To that end, we follow the taxonomy introduced in~\citet{raunak-etal-2021-curious} and later extended and studied in~\citet{guerreiro-etal-2023-looking}. Under this taxonomy, hallucinations are translations that contain content that is detached from the source sentence. To disentangle the different types of hallucinations, they can be categorized as: \textit{largely fluent detached hallucinations} or \textit{oscillatory hallucinations}. The former are translations that bear \textit{little or no relation at all} to the source content and may be further split according to the severity of the detachment (e.g. strong or full detachment) while the latter are inadequate translations that contain erroneous repetitions of words and phrases. We illustrate in Appendix~\ref{app:model_data_details} the categories described above through examples of hallucinated outputs. 

\section{On-the-fly detection of hallucinations}
On-the-fly hallucination detectors are systems that can detect hallucinations without access to reference translations. These detectors are particularly relevant as they can be deployed in online applications where references are not readily available.\footnote{As such, in this work, we will not consider metrics that depend on a reference sentence (e.g. chrF~\citep{popovic-2016-chrf}, COMET~\citep{rei-etal-2020-comet}). For an analysis on the performance of such metrics, please refer to~\citet{guerreiro-etal-2023-looking}.}

\subsection{Categorization of hallucination detectors}\label{subsec:ontheflydetection}
Previous work on on-the-fly detection of hallucinations in NMT has primarily focused on two categories of detectors: \textit{external} detectors and \textit{model-based} detectors. External detectors employ auxiliary models trained for related tasks such as quality estimation (QE) and cross-lingual embedding similarity. On the other hand, model-based detectors only require access to the NMT model that generates the translations, and work by leveraging relevant internal features such as model confidence and cross-attention. These detectors are attractive due to their flexibility and low memory footprint, as they can very easily be plugged in on a vast range of NMT models without the need for additional training data or computing infrastructure. Moreover,~\citet{guerreiro-etal-2023-looking} show that model-based detectors can be predictive of hallucinations, outperforming QE models and even performing on par with state-of-the-art reference-based metrics.

\subsection{Problem Statement}
We will focus specifically on model-based detectors that require obtaining internal features from a model $\mathcal{M}$. Building a hallucination detector generally consists of finding a scoring function $s_\mathcal{M} : \mathcal{X} \rightarrow \mathbb{R}$ and a threshold $\tau \in \mathbb{R}$ to build a binary rule $g_\mathcal{M} : \mathcal{X} \rightarrow \{0, 1\}$. For a given test sample $\bm{x} \in \mathcal{X}$, 
\begin{equation}
    g_\mathcal{M}(\bm{x}) = \mathbbm{1}{\{s_{\mathcal{M}}(\bm{x}) > \tau\}}. 
    \label{eq:detector}
\end{equation}
If $s_\mathcal{M}$ is an anomaly score, $g_\mathcal{M}(\bm{x})=0$ implies that the model $\mathcal{M}$ generates a `normal' translation for the source sequence $\bm{x}$, and $g_\mathcal{M}(\bm{x})=1$ implies that $\mathcal{M}$ generates a `hallucination' instead.\footnote{From now on, we will omit the subscript $\mathcal{M}$ from all model-based scoring functions to ease notation effort.}

\section{Unsupervised Hallucination Detection with Optimal Transport}
Anomalous cross-attention maps have been connected to the hallucinatory mode in several works~\citep{lee2018hallucinations, berard-etal-2019-naver, raunak-etal-2021-curious}. Our method builds on this idea and uses the Wasserstein distance to estimate the cost of transforming a translation source mass distribution into a reference distribution. Intuitively, the higher the cost of such transformation, the more distant---and hence the more anomalous--- the attention of the translation is with respect to that of the reference translation.

\begin{figure*}[t]
        \centering
        \includegraphics[width=\linewidth]{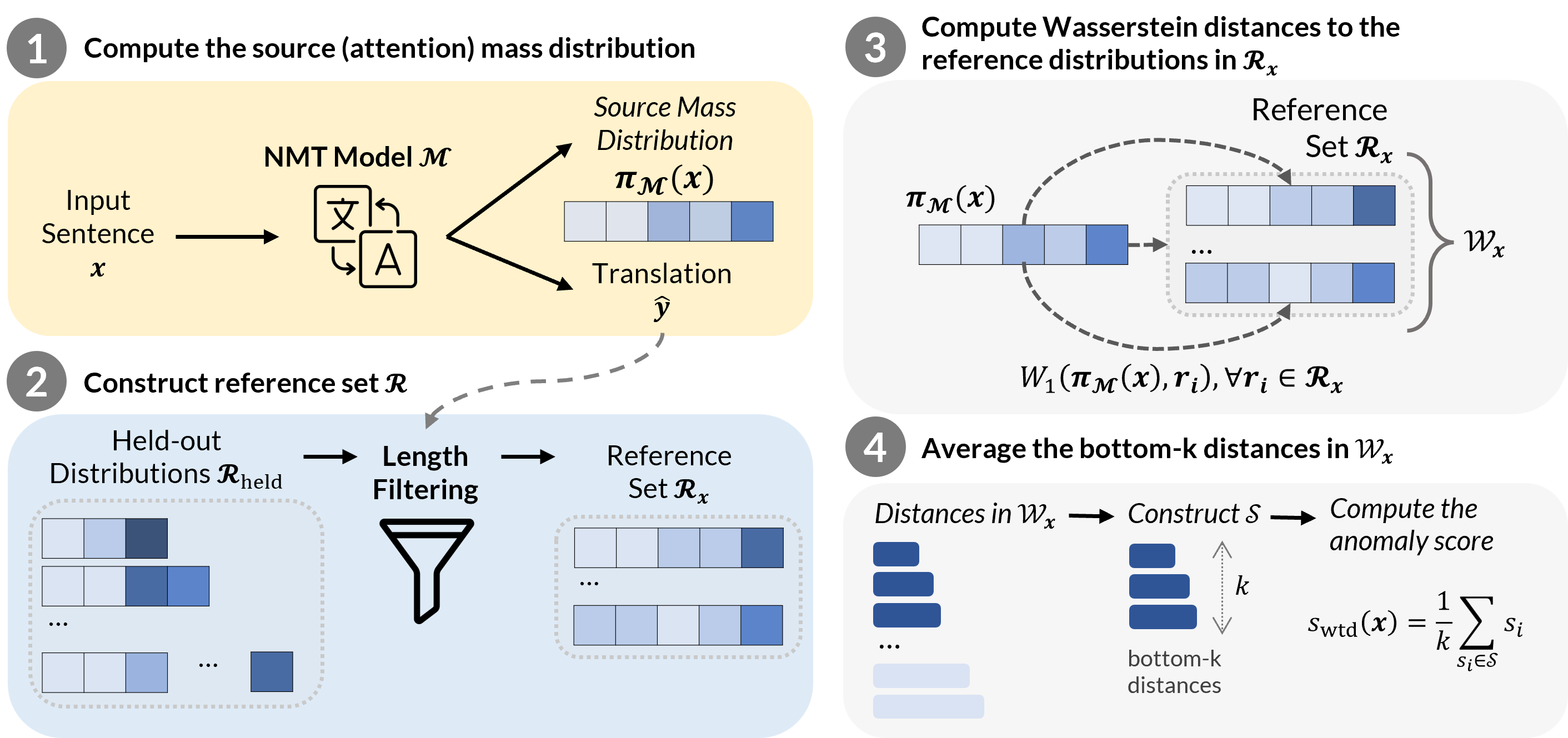}
    \caption{Procedure diagram for computation of the detection scores for the data-driven method \textsf{Wass-to-Data}.}
    \label{fig:diagramcomputation}
\end{figure*}

\subsection{\textsf{Wass-to-Unif}: A data independent scenario}
In this scenario, we only rely on the generated translation and its source mass distribution to decide whether the translation is a hallucination or not. Concretely, for a given test sample $\bm{x} \in \mathcal{X}$:
\begin{enumerate}[leftmargin=.45cm]
    \item We first obtain the source mass attention distribution $\bm{\pi}_{\mathcal{M}}(\bm{x})\in \triangle_{|\bm{x}|}$;
    \item We then compute an anomaly score, $ s_{\textsf{wtu}}(\bm{x})$, by measuring the Wasserstein distance between $\bm{\pi}_{\mathcal{M}}(\bm{x})$ and a reference distribution $\bm{u}$:
    \begin{equation}
    \label{eq:s_wtu}
    \textstyle s_{\textsf{wtu}}(\bm{x}) = W(\bm{\pi}_{\mathcal{M}}(\bm{x}), \bm{u}).
    \end{equation}
\end{enumerate}

\paragraph{Choice of reference translation.} A natural choice for $\bm{u}$ is the uniform distribution, $\bm{u} = \frac{1}{n} \cdot \bm{1}$, where $\bm{1}$ is a vector of ones of size $n$. In the context of our problem, a uniform source mass distribution means that all source tokens are equally attended.

\paragraph{Choice of cost function.} We consider the 0/1 cost function, $c(i,j) = \bm{1}[i\neq j]$, as it guarantees that the cost of transporting a unit mass from any token $i$ to any token $j \neq i$ is constant. For this distance function, the problem in Equation~\ref{eq:wass_dist} has the following closed-form solution~\citep{villani2009optimal}:
\begin{equation}\label{eq:wass_dist_uni}
W(\bm{\pi}_{\mathcal{M}}(\bm{x}), \boldsymbol{u}) = \nicefrac{1}{2}\, \|\bm{\pi}_{\mathcal{M}}(\bm{x})- \bm{u}\|_1. 
\end{equation}
This is a well-known result in optimal transport: the Wasserstein distance under the $0/1$ cost function is equivalent to the total variation distance between the two distributions. On this metric space, the Wasserstein distance depends solely on the probability mass that is transported to transform $\bm{\pi}_{\mathcal{M}}(\bm{x})$ to $\bm{u}$. Importantly, \textit{this formulation ignores the starting locations and destinations of that probability mass} as the cost of transporting a unit mass from any token $i$ to any token $j \neq i$ is constant.\vspace{5pt}

\noindent\textbf{Interpretation of \textsf{Wass-to-Unif}.\,} Attention maps for which the source attention mass is highly concentrated on a very sparse set of tokens (regardless of their location in the source sentence) can be very predictive of hallucinations~\citep{berard-etal-2019-naver, guerreiro-etal-2023-looking}. Thus, the bigger the distance between the source mass distribution of a test sample and the uniform distribution, the more peaked the former is, and hence the closer it is to such predictive patterns.

\subsection{\textsf{Wass-to-Data}: A data-driven scenario}
In this scenario, instead of using a single reference distribution, we use a set of reference source mass distributions, $\mathcal{R}_{\bm{x}}$, obtained with the same model. By doing so, we can evaluate how anomalous a given translation is compared to a model data-driven distribution, rather than relying on an arbitrary choice of reference distribution.

First, we use a held-out dataset $\mathcal{D}_{\textrm{held}}$ that contains samples for which the model $\mathcal{M}$ generates good quality translations according to an automatic evaluation metric (in this work, we use COMET~\citep{rei-etal-2020-comet}). We use this dataset to construct (offline) a set of held-out source attention distributions~$\mathcal{R}_{\textrm{held}} = \{\bm{\pi}_{\mathcal{M}}  (\bm{x}) \in \triangle_{|\bm{x}|}: \bm{x} \in \mathcal{D}_{\textrm{held}}\}$. Then, for a given test sample $\bm{x} \in \mathcal{X}$, we apply the procedure illustrated in Figure~\ref{fig:diagramcomputation}:
\begin{enumerate}[leftmargin=.45cm, itemsep=0em]
    \item We generate a translation $\hat{\bm{y}} = (y_1, \dots, y_m)$ and obtain the source mass attention distribution $\bm{\pi}_{\mathcal{M}}(\bm{x})\in \triangle_{|\bm{x}|}$;
    \item We apply a length filter to construct the sample reference set $\mathcal{R}_{\bm{x}}$, by restricting $\mathcal{R}_{\bm{x}}$ to contain source mass distributions of $\mathcal{R}_{\text{held}}$ correspondent to translations of size $[(1-\delta) m, (1+\delta) m]$ for a predefined $\delta \in ]0,1[$;\footnote{For efficiency reasons, we set the maximum cardinality of $\mathcal{R}_{\bm{x}}$ to $|\mathcal{R}|_{\textrm{max}}$. If length-filtering yields a set with more than $|\mathcal{R}|_{\textrm{max}}$ examples, we randomly sample $|\mathcal{R}|_{\textrm{max}}$ examples from that set to construct $\mathcal{R}_{\bm{x}}$.} 
    \item We compute pairwise Wasserstein-1 distances between $\bm{\pi}_{\mathcal{M}}(\bm{x})$ and each element $\bm{r}_i$ of $\mathcal{R}_{\bm{x}}$:
    \begin{align}\label{eq:set_W}
    \mathcal{W}_{\bm{x}}=\bigl(W_1(&\bm{\pi}_{\mathcal{M}}(\bm{x}), \bm{r}_1), \dots, \\ 
    & \left. W_1(\bm{\pi}_{\mathcal{M}}(\bm{x}), \bm{r}_{|\mathcal{R}_{\bm{x}}|})\right). \nonumber 
    \end{align}
    \item We obtain the anomaly score $ s_{\textsf{wtd}}(\bm{x})$ by  averaging the bottom-$k$ distances in $\mathcal{W}_{\bm{x}}$:
    \begin{equation}
       \textstyle s_{\textsf{wtd}}(\bm{x}) = \frac{1}{k}\, \sum_{s_i \in \mathcal{S}} s_i,
    \end{equation}
where $\mathcal{S}$ is the set containing the $k$ smallest elements of $\mathcal{W}_{\bm{x}}$. 

\end{enumerate}

\paragraph{Interpretation of \textsf{Wass-to-Data}.} Hallucinations, unlike good translations, are not fully supported by the source content. \textsf{Wass-to-Data} evaluates how anomalous a translation is by comparing the source attention mass distribution of that translation to those of good translations. The higher the \textsf{Wass-to-Data} score, the more anomalous the 
source attention mass distribution of that translation is in comparison to those of good translations, and the more likely it is to be an hallucination.

\paragraph{Relation to \textsf{Wass-to-Unif}.} The Wasserstein-1 distance~(see Section~\ref{subsec:wasserstein}) between two distributions is equivalent to the $\ell_1$-norm of the difference between their \textit{cumulative distribution functions}~\citep{COTFNT}. Note that this is different from the result in Equation~\ref{eq:wass_dist_uni}, as the Wasserstein distance under $c(i,j) = \bm{1}[i\neq j]$ as the cost function is proportional to the norm of the difference between their \textit{probability mass functions}. Thus, \textsf{Wass-to-Unif} will be more sensitive to the overall structure of the distributions (e.g. sharp probability peaks around some points), whereas \textsf{Wass-to-Data} will be more sensitive to the specific values of the points in the two distributions.


\subsection{\textsf{Wass-Combo}: The best of both worlds}
With this scoring function, we aim at combining \textsf{Wass-to-Unif} and \textsf{Wass-to-Data} into a single detector. To do so, we propose using a two-stage process that exploits the computational benefits of \textsf{Wass-to-Unif} over \textsf{Wass-to-Data}.\footnote{We also tested with a convex combination of the two detectors' scores. We present results for this alternative approach in Appendix~\ref{app:ablations}.} Put simply, (i)~we start by assessing whether a test sample is deemed a hallucination according to \textsf{Wass-to-Unif}, and if not (ii)~we compute the \textsf{Wass-to-Data} score. Formally,
\begin{align}\label{eq:wass_combo}
    s_{\textsf{wc}}(\bm{x}) = \mathbbm{1} &\left[s_{\textsf{wtu}}(\bm{x}) > \tau_{\textsf{wtu}} \right] \times \tilde{s}_{\textsf{wtu}}(\bm{x}) \\ &+ \mathbbm{1} \left[s_{\textsf{wtu}}(\bm{x}) \leq \tau_{\textsf{wtu}} \right] \times s_{\textsf{wtd}}(\bm{x}) \nonumber
\end{align}
for a predefined scalar threshold $\tau_{\textsf{wtu}}$. To set that threshold, we compute~$\mathcal{W}_{\textsf{wtu}} = \{s_{\textsf{wtu}} (\bm{x}): \bm{x} \in \mathcal{D}_{\textrm{held}}\}$ and set $\tau_{\textsf{wtu}}=P_{K}$, i.e $\tau_{\textsf{wtu}}$ is the $K$\textsuperscript{th} percentile of $\mathcal{W}_{\textsf{wtu}}$ with $K \in\, ]98,100[$ (in line with hallucinatory rates reported in~\citep{muller-etal-2020-domain, wang-sennrich-2020-exposure, salted_raunak2022}).\footnote{In order to make the scales of ${s}_{\textsf{wtu}}$ and ${s}_{\textsf{wtd}}$ compatible, we use a scaled $\tilde{s}_{\textsf{wtu}}$ value instead of ${s}_{\textsf{wtu}}$ in Equation~\ref{eq:wass_combo}. We obtain $\tilde{s}_{\textsf{wtu}}$ by min-max scaling $s_{\textsf{wtu}}$ such that $\tilde{s}_{\textsf{wtu}}$ is within the range of ${s}_{\textsf{wtd}}$ values obtained for a held-out set.}

\section{Experimental Setup}
\subsection{Model and Data} 
We follow the setup in~\citet{guerreiro-etal-2023-looking}. In that work, the authors released a dataset of 3415 translations for WMT18 \textsc{de-en} news translation data~\citep{bojar-etal-2018-findings} with annotations on critical errors and hallucinations. Our analysis in the main text focuses on this dataset as it is the only available dataset that contains human annotations on hallucinations produced naturally by an NMT model (we provide full details about the dataset and the model in Appendix~\ref{app:model_data_details}). Nevertheless, in order to access the broader validity of our methods for other low and mid-resource language pairs and models, we follow a similar setup to that of~\citet{tangspeciahallreduction2022} in which quality assessments are converted to hallucination annotations. For those experiments, we use the \textsc{ro-en}~(mid-resource) and \textsc{ne-en}~(low-resource) translations from the MLQE-PE dataset~\citep{fomicheva-etal-2022-mlqe}. In Appendix~\ref{app:mlqepe_experiments}, we present full details on the setup and report the results of these experiments. Importantly, our empirical observations are similar to those of the main text. For all our experiments, we obtain all model-based information required to build the detectors using the same models that generated the translations in consideration.

\begin{figure*}[t]
    \centering
    \includegraphics[width=\textwidth]{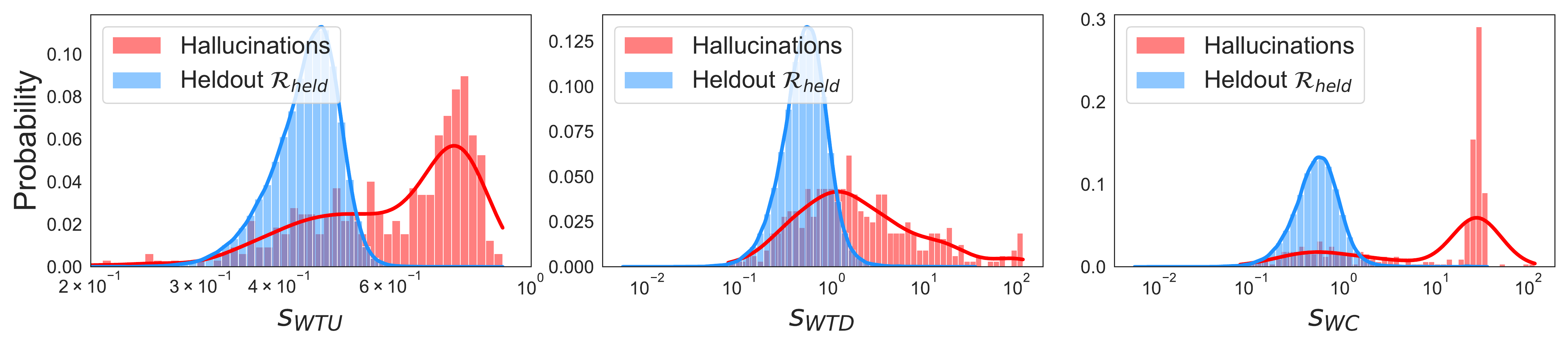}
    \caption{Histogram scores for our methods -- \textsf{Wass-to-Unif} (left), \textsf{Wass-to-Data} (center) and \textsf{Wass-Combo} (right). We display \textsf{Wass-to-Data} and \textsf{Wass-Combo} scores on log-scale.}
    \label{fig:histograms_wassdist}
\end{figure*}

\subsection{Baseline detectors}
\subsubsection{Model-based detectors}
We compare our methods to the two best performing model-based methods in~\citet{guerreiro-etal-2023-looking}.\footnote{We compare with ALTI+~\citep{altijavier2022}, a method thas was leveraged for hallucination detection concurrently to our work in~\citet{dale2022detecting}, in Appendix~\ref{app:alti}.}

\paragraph{\textsf{Attn-ign-SRC}.} This method consists of computing the proportion of source words with a total incoming attention mass lower than a threshold $\lambda$:
\begin{equation}
    \textstyle s_{\textsf{ais}}(\bm{x}) = \frac{1}{n} \sum_{j=1}^n \mathbbm{1}\left[(\bm{\Omega}^\top(\bm{x})\bm{1})_j < \lambda \right].
    \label{eq:attnignsrc}  
\end{equation}
This method was initially proposed in \citet{berard-etal-2019-naver}. We follow their work and use $\lambda = 0.2$.

\paragraph{\textsf{Seq-Logprob.}} We compute the length-normalised sequence log-probability of the translation:
\begin{equation}
    \textstyle s_{\textsf{slp}}(\bm{x}) = \frac{1}{m} \sum_{k=1}^{m} \log p_{\bm{\theta}}(y_{k} \mid \bm{y}_{<k}, \bm{x}).
    \label{eq:seq_logprob}
\end{equation}

\subsubsection{External detectors}\label{subsubsec:external}

We provide a comparison to detectors that exploit state-of-the-art models in related tasks, as it helps monitor the development of model-based detectors.

\paragraph{\textsf{CometKiwi}.} We compute sentence-level quality scores with \textsf{CometKiwi}~\citep{cometkiwi2022}, the winning reference-free model of the WMT22 QE shared task~\citep{zerva-EtAl:2022:WMT}. It has more than 565M parameters and it was trained on more than 1M human quality annotations. Importantly, this training data includes human annotations for several low-quality translations and hallucinations.

\paragraph{\textsf{LaBSE}.} We leverage \textsf{LaBSE}~\citep{labse2022} to compute cross-lingual sentence representations for the source sequence and translation. We use the cosine similarity of these representations as the detection score. The model is based on the BERT~\citep{devlin-etal-2019-bert} architecture and was trained on more than 20 billion sentences. \textsf{LaBSE} makes for a good baseline, as it was optimized in a self-supervised way with a translate matching objective that is very much aligned with the task of hallucination detection: during training, \textsf{LaBSE} is given a source sequence and a set of translations including the true translation and multiple negative alternatives, and the model is optimized to specifically discriminate the true translation from the other negative alternatives by assigning a higher similarity score to the former. 

\subsection{Evaluation metrics} 
We report the Area Under the Receiver Operating Characteristic curve (AUROC) and the False Positive Rate at 90\% True Positive Rate (FPR@90TPR) to evaluate the performance of different detectors.

\subsection{Implementation Details}\label{subsec:implementation_details}
We use WMT18 \textsc{de-en} data samples from the held-out set used in~\citet{guerreiro-etal-2023-looking}, and construct $\mathcal{D}_{\text{held}}$ to contain the 250k samples with highest COMET score. To obtain \textsf{Wass-to-Data} scores, we set $\delta = 0.1$, $|\mathcal{R}|_{\text{max}} = 1000$ and $k=4$. To obtain \textsf{Wass-to-Combo} scores, we set $\tau_{\textsf{wtu}}=P_{99.9}$. We perform extensive ablations on the construction of $\mathcal{R}_{\text{held}}$ and on all other hyperparameters in Appendix~\ref{app:ablations}. We also report the computational runtime of our methods in Appendix~\ref{app:computationalruntime}.

\section{Results}
\label{sec:results}
\subsection{Performance on on-the-fly detection}\label{sec:onthefly_analysis}

We start by analyzing the performance of our proposed detectors on a real world on-the-fly detection scenario. In this scenario, the detector must be able to flag hallucinations \textit{regardless of their specific type} as those are unknown at the time of detection.


\definecolor{RoyalBlue}{HTML}{00a572}
\begin{table}[t]
\centering
\renewcommand\arraystretch{1}
\footnotesize
\begin{tabular}{>{\arraybackslash}m{2.05cm} r r}
\toprule
\textsc{Detector} & AUROC $\uparrow$ & FPR@90TPR $\downarrow$\\
\midrule 
\multicolumn{3}{l}{\textbf{\textit{External Detectors}}} \\
{\textsf{CometKiwi}} & {86.96} & 53.61
\\
\textsf{LaBSE} & \textbf{91.72} & \textbf{26.91} \\\cdashlinelr{1-3}\noalign{\vskip 0.5ex} 

\multicolumn{3}{l}{\textbf{\textit{Model-based Detectors}}} \\
\textsf{Attn-ign-SRC}   & 79.36 & 72.83 \\
\textsf{Seq-Logprob}    & 83.40 & 59.02\\[0.1cm]
\rowcolor{gray!8} \multicolumn{3}{l}{\textbf{\textsc{Ours}}}\\\noalign{\vskip 0.25ex}
\textsf{Wass-to-Unif} & 80.37 & 72.22 \\
\textsf{Wass-to-Data} & 84.20$_{\scriptsize{\textcolor{gray}{\, 0.15}}}$ & \textbf{48.15}$_{\scriptsize{\textcolor{gray}{\, 0.54}}}$ \\
\textsf{Wass-Combo} & \textbf{87.17}$_{\scriptsize{\textcolor{gray}{\, 0.07}}}$ & \textbf{47.56}$_{\scriptsize{\textcolor{gray}{\, 1.30}}}$\\
\bottomrule
\end{tabular}
\caption{Performance of all hallucination detectors. For \textsf{Wass-to-Data} and \textsf{Wass-Combo} we present the mean and standard deviation scores across five random seeds.}
\label{tab:auroc_fpr_all}
\end{table}

\paragraph{\textsf{Wass-Combo} is the best model-based detector.} Table~\ref{tab:auroc_fpr_all} shows that \textsf{Wass-Combo} outperforms most other methods both in terms of AUROC and FPR. When compared to the previous best-performing model-based method (\textsf{Seq-Logprob}), \textsf{Wass-Combo} obtains boosts of approximately $4$ and $10$ points in AUROC and FPR, respectively. These performance boosts are further evidence that model-based features can be leveraged, in an unsupervised manner, to build effective detectors. Nevertheless, the high values of FPR suggest that there is still a significant performance margin to reduce in future research.

\paragraph{The notion of data proximity is helpful to detect hallucinations.} Table~\ref{tab:auroc_fpr_all} shows that \textsf{Wass-to-Data} outperforms the previous best-performing model-based method (\textsf{Seq-Logprob}) in both AUROC and FPR (by more than $10\%$). This supports the idea that cross-attention patterns for hallucinations are anomalous with respect to those of good model-generated translations, and that our method can effectively measure this level of anomalousness. On the other hand, compared to \textsf{Wass-to-Uni}, \textsf{Wass-to-Data} shows a significant improvement of 30 FPR points. This highlights the effectiveness of leveraging the data-driven distribution of good translations instead of the ad-hoc uniform distribution. Nevertheless, Table~\ref{tab:auroc_fpr_all} and Figure~\ref{fig:histograms_wassdist} show that combining both methods brings further performance improvements. This suggests that these methods may specialize in different types of hallucinations, and that combining them allows for detecting a broader range of anomalies. We will analyze this further in Section~\ref{sec:type_analysis}.

\paragraph{Our model-based method achieves comparable performance to external models.} Table~\ref{tab:auroc_fpr_all} shows that \textsf{Wass-Combo} outperforms \textsf{CometKiwi}, with significant improvements on FPR. However, there still exists a gap to \textsf{LaBSE}, the best overall detector. This performance gap indicates that more powerful detectors can be built, paving the way for future work in model-based hallucination detection. Nevertheless, while relying on external models seems appealing, deploying and serving them in practice usually comes with additional infrastructure costs, while our detector relies on information that can be obtained when generating the translation.

\paragraph{Translation quality assessments are less predictive than similarity of cross-lingual sentence representations.} Table~\ref{tab:auroc_fpr_all} shows that \textsf{LaBSE} outperforms the state-of-the-art quality estimation system \textsf{CometKiwi}, with vast improvements in terms of FPR. This shows that for hallucination detection, quality assessments obtained with a QE model are less predictive than the similarity between cross-lingual sentence representations. This may be explained through their training objectives~(see Section~\ref{subsubsec:external}): while \textsf{CometKiwi} employs a more general regression objective in which the model is trained to match human quality assessments, \textsf{LaBSE} is trained with a translate matching training objective that is very closely related to the task of hallucination detection.

\begin{figure}[t]
    \centering
    \includegraphics[width=\linewidth]{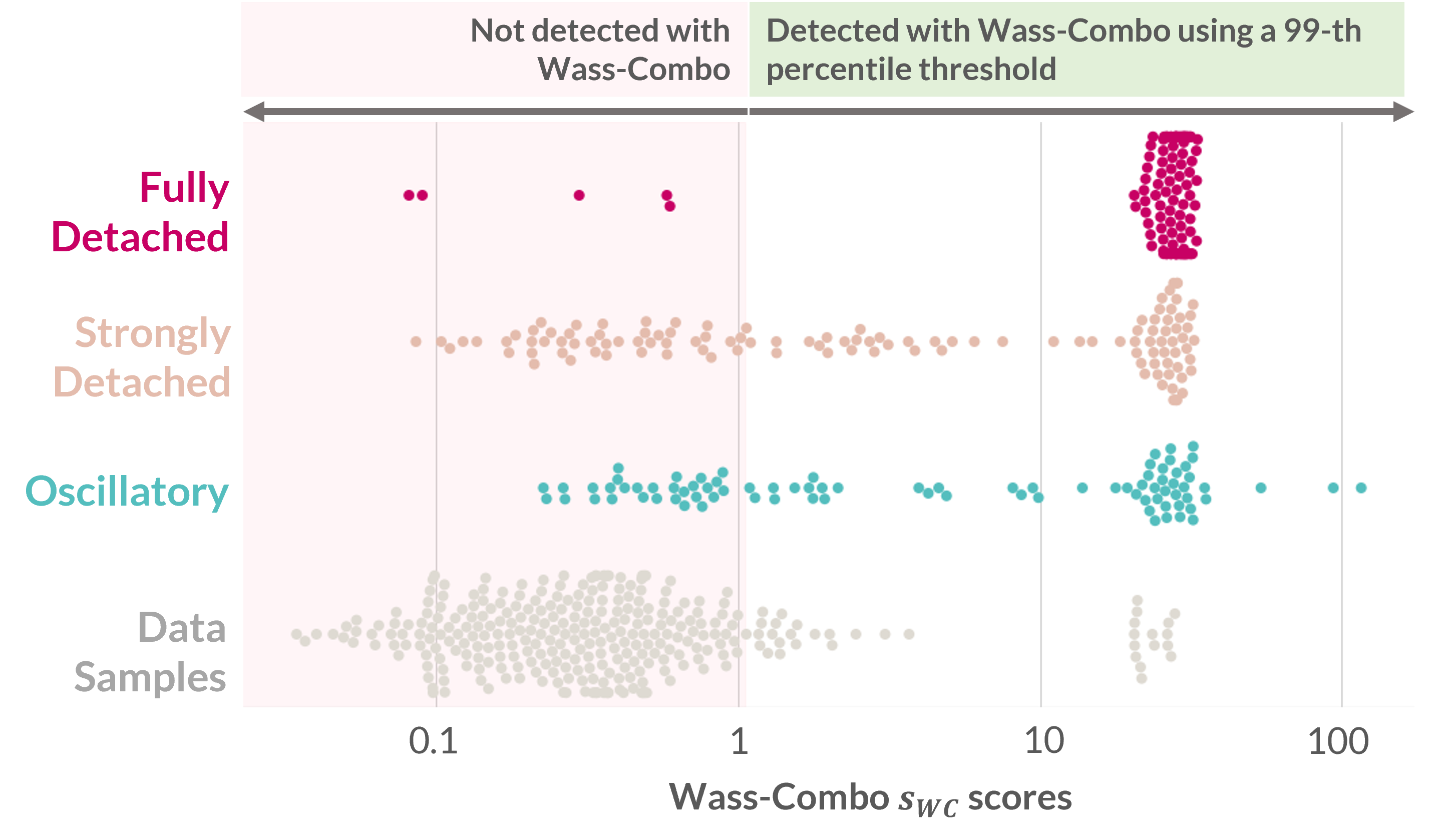}
    \caption{Distribution of \textsf{Wass-Combo} scores (on log-scale) for hallucinations and data samples, and performance on a fixed-threshold scenario.}
    \label{fig:errors_halls}
\end{figure}

\begin{table*}[t]
\centering
\footnotesize
\renewcommand\arraystretch{1}
\setlength{\tabcolsep}{4.5pt}
\begin{subtable}[t]{0.45\linewidth}
\centering
\begin{tabular}{>{\arraybackslash}m{1.9cm} r r r}
\toprule
\multirow{ 2}{*}{\textsc{Detector}}  & \textbf{Fully} & \multirow{ 2}{*}{{\textbf{Oscillatory}}} & {\textbf{Strongly}} \\
& \textbf{Detached} & & \textbf{Detached} \\
\midrule
\multicolumn{4}{l}{\textbf{\textit{External Detectors}}} \\
{\textsf{CometKiwi}} & {87.75} & {\textbf{93.04}} & 81.78 \\
{\textsf{LaBSE}} & \textbf{98.91} &  84.62   & {\textbf{89.72}}\\\cdashlinelr{1-4}\noalign{\vskip 0.5ex} 
\multicolumn{4}{l}{\textbf{\textit{Model-based Detectors}}} \\
\textsf{Attn-ign-SRC}   & 95.76 & 59.53 & 77.42 \\
\textsf{Seq-Logprob}    & 95.64 & 71.10 & \textbf{80.15}\\[0.075cm]
\rowcolor{gray!8} \multicolumn{4}{l}{\textbf{\textsc{Ours}}}\\\noalign{\vskip 0.25ex} 
\textsf{Wass-to-Unif} & 96.35 & 69.75 & 72.19 \\
\textsf{Wass-to-Data} & 88.24$_{\scriptsize{\textcolor{gray}{\, 0.29}}}$ & \textbf{87.80}$_{\scriptsize{\textcolor{gray}{\, 0.10}}}$ & 77.60$_{\scriptsize{\textcolor{gray}{\, 0.18}}}$ \\
\textsf{Wass-Combo} & \textbf{96.57}$_{\scriptsize{\textcolor{gray}{\, 0.10}}}$ & 85.74$_{\scriptsize{\textcolor{gray}{\, 0.10}}}$ & 78.89$_{\scriptsize{\textcolor{gray}{\, 0.15}}}$\\
\bottomrule
\end{tabular}
\caption{AUROC -- the higher the better.}
\end{subtable}
\hfill
\begin{subtable}[t]{0.45\linewidth}
\centering
\begin{tabular}{>{\arraybackslash}m{1.9cm} r r r}
\toprule
\multirow{ 2}{*}{\textsc{Detector}}  & \textbf{Fully} & \multirow{ 2}{*}{{\textbf{Oscillatory}}} & {\textbf{Strongly}} \\
& \textbf{Detached} & & \textbf{Detached} \\
\midrule
\multicolumn{4}{l}{\textbf{\textit{External Detectors}}} \\
{\textsf{CometKiwi}} & {33.70} & {\textbf{23.80}} & 42.98 \\
{\textsf{LaBSE}} & \textbf{0.52} & 50.26 & \textbf{28.88} \\\cdashlinelr{1-4}\noalign{\vskip 0.5ex} 
\multicolumn{4}{l}{\textbf{\textit{Model-based Detectors}}} \\
\textsf{Attn-ign-SRC}   & 8.51 & 81.24 & 76.68 \\
\textsf{Seq-Logprob}    & 4.62 & 72.99 & \textbf{65.39}\\[0.075cm]
\rowcolor{gray!8} \multicolumn{4}{l}{\textbf{\textsc{Ours}}}\\\noalign{\vskip 0.25ex}
\textsf{Wass-to-Unif} & \textbf{3.27} & 78.78 & 88.32 \\
\textsf{Wass-to-Data} & 36.60$_{\scriptsize{\textcolor{gray}{\, 1.92}}}$ & \textbf{40.04}$_{\scriptsize{\textcolor{gray}{\, 1.57}}}$ & \textbf{63.96}$_{\scriptsize{\textcolor{gray}{\, 2.04}}}$ \\
\textsf{Wass-Combo} & 3.56$_{\scriptsize{\textcolor{gray}{\, 0.00}}}$ & \textbf{41.38}$_{\scriptsize{\textcolor{gray}{\, 1.59}}}$ & \textbf{64.55}$_{\scriptsize{\textcolor{gray}{\, 1.93}}}$ \\
\bottomrule
\end{tabular}
\centering
\caption{FPR@90TPR ($\%$) -- the lower the better.}
\end{subtable} \hspace{10pt}
\caption{Performance of all hallucination detectors for each hallucination type. For \textsf{Wass-to-Data} and \textsf{Wass-Combo}, we present the mean and standard deviation across five random seeds.}
\label{tab:auroc_fpr_type}
\end{table*}

\subsection{Do detectors specialize in different types of hallucinations?} \label{sec:type_analysis} 
In this section, we present an analysis on the performance of different detectors for different types of hallucinations (see Section~\ref{ssec:hallucinationsinnmt}). We report both a quantitative analysis to understand whether a detector can distinguish a specific hallucination type from other translations (Table~\ref{tab:auroc_fpr_type}), and a qualitative analysis on a fixed-threshold scenario\footnote{We set the threshold by finding the 99\textsuperscript{th} percentile of \textsf{Wass-Combo} scores obtained for 100k samples from the clean WMT18 \textsc{de-en} held-out set (see Section~\ref{subsec:implementation_details}).} (Figure~\ref{fig:errors_halls}). This analysis is particularly relevant to better understand how different detectors specialize in different types of hallucinations. In Appendix~\ref{app:mlqepe_experiments}, we show that the trends presented in this section hold for other mid- and low-resource language pairs.

\paragraph{Fully detached hallucinations.} Detecting fully detached hallucinations is remarkably easy for most detectors. Interestingly, \textsf{Wass-to-Unif} significantly outperforms \textsf{Wass-to-Data} on this type of hallucination. This highlights how combining both methods can be helpful. In fact, \textsf{Wass-Combo} performs similarly to \textsf{Wass-to-Unif}, and can very easily separate most fully detached hallucinations from other translations on a fixed-threshold scenario~(Figure~\ref{fig:errors_halls}). Note that the performance of \textsf{Wass-to-Unif} for fully detached hallucinations closely mirrors that of \textsf{Attn-ign-SRC}. This is not surprising, since both methods, at their core, try to capture similar patterns: translations for which the source attention mass distribution is highly concentrated on a small set of source tokens.

\paragraph{Strongly detached hallucinations.} These are the hardest hallucinations to detect with our methods. Nevertheless, \textsf{Wass-Combo} performs competitively with the previous best-performing model-based method for this type of hallucinations (\textsf{Seq-Logprob}). We hypothesize that the difficulty in detecting these hallucinations may be due to the varying level of detachment from the source sequence. Indeed, Figure~\ref{fig:errors_halls} shows that \textsf{Wass-Combo} scores span from a cluster of strongly detached hallucinations with scores similar to other data samples to those similar to the scores of most fully detached hallucinations.

\paragraph{Oscillatory hallucinations.} \textsf{Wass-to-Data} and \textsf{Wass-Combo} significantly outperform all previous model-based detectors on detecting oscillatory hallucinations. This is relevance in the context of model-based detectors, as previous detectors notably struggle with detecting these hallucinations. Moreover, \textsf{Wass-Combo} also manages to outperform \textsf{LaBSE} with significant improvements in FPR. This hints that the repetition of words or phrases may not be enough to create sentence-level representations that are highly dissimilar from the non-oscillatory source sequence. In contrast, we find that \textsf{CometKiwi} appropriately penalizes oscillatory hallucinations, which aligns with observations made in~\citet{guerreiro-etal-2023-looking}.  

Additionally, Figure~\ref{fig:errors_halls} shows that the scores for oscillatory hallucinations are scattered along a broad range. After close evaluation, we observed that this is highly related to the severity of the oscillation: almost all non-detected hallucinations are not severe oscillations~(see Appendix~\ref{app:error_analysis}).

\section{Conclusions}
We propose a novel plug-in model-based detector for hallucinations in NMT. Unlike previous attempts to build an attention-based detector, we do not rely on \textit{ad-hoc} heuristics to detect hallucinations, and instead pose hallucination detection as an optimal transport problem: our detector aims to find translations whose source attention mass distribution is highly distant from those of good quality translations. Our empirical analysis shows that our detector outperforms all previous model-based detectors. Importantly, in contrast to these prior approaches, it is suitable for identifying oscillatory hallucinations, thus addressing an important gap in the field. We also show that our detector is competitive with external detectors that use state-of-the-art quality estimation or cross-lingual similarity models. Notably, this performance is achieved without the need for large models, or any data with quality annotations or parallel training data.  Finally, thanks to its flexibility, our detector can be easily deployed in real-world scenarios, making it a valuable tool for practical applications.

\section*{Limitations}
We highlight two main limitations of our work. 

Firstly, instead of focusing on more recent NMT models that use large pretrained language models as their backbone, our experiments were based on transformer base models. That is because we used the NMT models that produced the translations in the datasets we analyze, i.e, the models that actually \textit{hallucinate} for the source sequences in the dataset. Nevertheless, research on hallucinations for larger NMT models makes for an exciting line of future work and would be valuable to assess the broad validity of our claims.

Secondly, although our method does not require any training data or human annotations, it relies on access to a pre-existing database of source mass distributions. This can be easily obtained offline by running the model on monolingual data to obtain the distributions. Nevertheless, these datastores need not be costly in terms of memory. In fact, in Appendix~\ref{app:mlqepe_experiments}, we validate our detectors for datastores that contain less than 100k distributions.


\section*{Acknowledgments}
This work is partially supported by the European Research Council (ERC StG DeepSPIN 758969), by EU’s
Horizon Europe Research and Innovation Actions (UTTER, contract 101070631), by the P2020 program MAIA (LISBOA-01-0247-FEDER-045909), by the Portuguese Recovery and Resilience Plan  through project C645008882-00000055 (NextGenAI, Center for Responsible AI), and by the FCT through contract UIDB/50008/2020. This work was also granted access to the HPC resources of IDRIS under the allocation 2021- AP010611665 as well as under the project 2021- 101838 made by GENCI.

\bibliography{anthology,custom}
\bibliographystyle{acl_natbib}
\clearpage 
\appendix 
\input{appendix}
\end{document}

%% file: math_commands.tex
\usepackage{amsmath,amsfonts,bm}

\def\eqref#1{equation~\ref{#1}}

\def\1{\bm{1}}

\DeclareMathAlphabet{\mathsfit}{\encodingdefault}{\sfdefault}{m}{sl}
\SetMathAlphabet{\mathsfit}{bold}{\encodingdefault}{\sfdefault}{bx}{n}



%% file: appendix.tex
\section{Model and Data Details}
\label{app:model_data_details}

\paragraph{NMT Model.} The NMT model used in \citet{guerreiroetal2022} to create the hallucination dataset is a Transformer base model~\cite{transformer_vaswani}  (hidden size of 512, feedforward size of 2048, 6 encoder and 6 decoder layers, 8 attention heads). The model has approximately 77M parameters. It was trained with the \texttt{fairseq} toolkit~\citep{ott-etal-2019-fairseq} on WMT18 \textsc{de-en} data (excluding Paracrawl): the authors randomly choose 2/3 of the dataset for training and use the remaining 1/3 as a held-out set for analysis. We use that same held-out set in this work. 

\paragraph{Dataset Stats.} The dataset used in this paper was introduced in~\citet{guerreiroetal2022}. It consists of 3415 translations from WMT18 \textsc{de-en} data with structured annotations on different types of hallucinations and pathologies. Overall, the dataset contains 118 translations annotated as fully detached hallucinations, 90 as strongly detached hallucinations, and 86 as oscillatory hallucinations.\footnote{Some strongly detached hallucinations have also been annotated as oscillatory hallucinations. In these cases, we consider them to be oscillatory.} The other translations are either incorrect (1073) or correct (2048). Details on annotation, a high-level overview and other statistics can be found in the original paper. We show examples of hallucinations for each category in Table~\ref{tab:hallucinationexamples}.\footnote{All data used in this paper is licensed under a MIT License.}


\begin{table*}[t]
\centering
\scriptsize
\renewcommand\arraystretch{1.0}
\begin{tabular}{
>{\arraybackslash}m{1.0cm} >{\arraybackslash}m{4.8cm}
>{\arraybackslash}m{4.5cm}
>{\arraybackslash}m{4cm}}
\toprule
\textbf{Category} &  \textbf{Source Sentence} & \textbf{Reference Translation} & \textbf{Hallucination}\\ \midrule
Oscillatory & Als Maß hierfür wird meist der sogenannte Pearl Index benutzt (so benannt nach einem Statistiker, der diese Berechnungsformel einführte). & As a measure of this, the so-called Pearl Index is usually used (so named after a statistician who introduced this calculation formula). & The \textcolor{HallRedText}{\hl{term "Pearl Index"}} refers to \textcolor{HallRedText}{\hl{the term "Pearl Index"}} (or \textcolor{HallRedText}{\hl{"Pearl Index"}}) used to refer to \textcolor{HallRedText}{\hl{the term "Pearl Index"}} (or \textcolor{HallRedText}{\hl{"Pearl Index"}}).\\[1.25ex] \midrule
Strongly Detached & Fraktion der Grünen / Freie Europäische Allianz & The Group of the Greens/European Free Alliance & \textcolor{HallRedText}{\hl{Independence and Democracy}} Group \textcolor{HallRedText}{\hl{(includes 10 UKIP MEPs and one independent MEP from Ireland)}} \\[1.25ex] \midrule
Fully \ \ \ \ \ \  Detached & Die Zimmer beziehen, die Fenster mit Aussicht öffnen, tief durchatmen, staunen. & Head up to the rooms, open up the windows and savour the view, breathe deeply, marvel. &  \textcolor{HallRedText}{\hl{The staff were very friendly and helpful.}} \\ 
\bottomrule
\end{tabular}
\caption{Examples of hallucination types. Hallucinated content is shown \textcolor{HallRedText}{\hl{shaded}}.}
\label{tab:hallucinationexamples}
\end{table*}


\section{Details on External Detectors}

\paragraph{\textsf{COMET}.} We use models available in the official repository\footnote{\url{https://github.com/Unbabel/COMET}}: \texttt{wmt22-cometkiwi-da} for CometKiwi and \texttt{wmt20-comet-da} for COMET.

\paragraph{\textsf{LaBSE}.} We use the version available in \texttt{sentence-transformers}~\citep{reimers-2019-sentence-bert}.\footnote{\url{https://huggingface.co/sentence-transformers/LaBSE}}

\section{Performance of reference-free COMET-based models}
\label{app:cometperformance}
\citet{guerreiroetal2022} used the COMET-QE version \texttt{wmt20-comet-qe-da}, whereas we are using the latest iteration \texttt{wmt22-cometkiwi-da} (CometKiwi). CometKiwi was trained on human annotations from the MLQE-PE dataset~\citep{fomicheva-etal-2022-mlqe}, which contains a high percentage of hallucinations for some language pairs~\citep{specia-etal-2021-findings, tangspeciahallreduction2022}. We show the performance of both these versions in Table~\ref{tab:comet_version_performance}. CometKiwi significantly outperforms the previous iteration of COMET-QE. This hints that training quality estimation models with more negative examples can improve their ability to adequately penalize hallucinations.

\begin{table}[t]
\centering
\renewcommand\arraystretch{1.0}
\footnotesize
\begin{tabular}{>{\arraybackslash}m{2.45cm} r r}
\toprule
\textsc{Model Version} & AUROC $\uparrow$ & FPR@90TPR $\downarrow$\\
\midrule 
\texttt{wmt20-comet-qe-da} & 70.15 & 57.24 \\
\texttt{wmt22-cometkiwi-da} & \textbf{86.96} & \textbf{53.61} \\
\bottomrule
\end{tabular}
\caption{Performance of COMET-QE (\texttt{wmt20-comet-qe-da}) and CometKiwi (\texttt{wmt22-cometkiwi-da}) on the on-the-fly detection scenario.}
\label{tab:comet_version_performance}
\end{table}

\section{Computational runtime of our detectors}
\label{app:computationalruntime}
Our detectors do not require access to a GPU machine. All our experiments have been ran on a machine with 2 physical Intel(R) Xeon(R) Gold 6348 @ 2.60GHz CPUs (total of 112 threads). Obtaining \textsf{Wass-to-Unif} scores for all the 3415 translations from the~\citet{guerreiroetal2022} dataset takes less than half a second, while \textsf{Wass-to-Data} scores are obtained in little over 4 minutes.

\section{Evaluation Metrics}
\label{app:eval_metrics}
We use \texttt{scikit-learn}~\citep{scikit-learn} implementations of our evaluation metrics.\footnote{\url{https://scikit-learn.org
}}

\section{Tracing-back performance boosts to the construction of the reference set~$\mathcal{R}_{\bm{x}}$}\label{subsec:ablations}
In Section~\ref{sec:onthefly_analysis} in the main text, we showed that evaluating how distant a given translation is compared to a data-driven reference distribution--rather than to an \textit{ad-hoc} reference distribution-- led to increased performance. Therefore, we will now analyze the construction of the reference set $\mathcal{R}_{\bm{x}}$ to obtain \textsf{Wass-to-Data} scores (step 2 in Figure~\ref{fig:diagramcomputation}). We conduct experiments to investigate the importance of the two main operations in this process: defining and length-filtering the distributions in $\mathcal{R}_{\text{held}}$.

\begin{table}[t]
\centering
\renewcommand\arraystretch{1.0}
\footnotesize
\begin{tabular}{>{\arraybackslash}m{2.05cm} r r}
\toprule
\textsc{Ablation} & AUROC $\uparrow$ & FPR@90TPR $\downarrow$\\
\midrule 
\multicolumn{3}{l}{\textbf{\textit{Model-Generated Translations}}} \\
Any & 83.27$_{\scriptsize\textcolor{gray}{\, 0.39}}$ & 50.08$_{\scriptsize\textcolor{gray}{\, 1.65}}$
\\
Quality-filtered & \textbf{84.20}$_{\scriptsize\textcolor{gray}{\, 0.15}}$ & \textbf{48.15}$_{\scriptsize\textcolor{gray}{\, 0.54}}$
\\\cdashlinelr{1-3}\noalign{\vskip 0.5ex}   
\multicolumn{3}{l}{\textbf{\textit{Reference Translations}}} \\
Any & \textbf{83.95}$_{\scriptsize\textcolor{gray}{\, 0.16}}$ & 50.26$_{\scriptsize\textcolor{gray}{\, 0.60}}$ \\
\bottomrule
\end{tabular}
\caption{Ablations on \textsf{Wass-to-Data} by changing the construction of $\mathcal{R}_{\text{held}}$. We present the mean and standard deviation (in subscript) across five random seeds.}
\label{tab:ablation_rheld}
\end{table}

\begin{table}[t]
\centering
\renewcommand\arraystretch{1.0}
\footnotesize
\begin{tabular}{>{\arraybackslash}m{2.45cm} r r}
\toprule
\textsc{Ablation} & AUROC $\uparrow$ & FPR@90TPR $\downarrow$\\
\midrule 
Random Sampling & 80.65$_{\scriptsize\textcolor{gray}{\, 0.15}}$ & 57.06$_{\scriptsize\textcolor{gray}{\, 2.04}}$
\\
Length Filtering & \textbf{84.20}$_{\scriptsize\textcolor{gray}{\, 0.15}}$ & \textbf{48.15}$_{\scriptsize\textcolor{gray}{\, 0.54}}$ \\
\bottomrule
\end{tabular}
\caption{Ablations on \textsf{Wass-to-Data} by changing the length-filtering window to construct $\mathcal{R}_{\bm{x}}$. We present the mean and standard deviation (in subscript) across five random seeds.}
\label{tab:ablation_lenfilter}
\end{table}


\paragraph{Construction of $\mathcal{R}_{\text{held}}$.} To construct $\mathcal{R}_{\text{held}}$, we first need to obtain the source attention mass distributions for each sample in $\mathcal{D}_{\text{held}}$. If $\mathcal{D}_{\text{held}}$ is a parallel corpus, we can force-decode the reference translations to construct $\mathcal{R}_{\text{held}}$. As shown in Table~\ref{tab:ablation_rheld}, this construction produces results similar to using good-quality model-generated translations. Moreover, we also evaluate the scenario where $\mathcal{R}_{\text{held}}$ is constructed with translations of any quality. Table~\ref{tab:ablation_rheld} shows that although filtering for quality improves performance, the gains are not substantial. This connects to findings by~\citet{guerreiroetal2022}: hallucinations exhibit different properties from other translations, including other incorrect translations. We offer further evidence that properties of hallucinations---in this case, the source attention mass distributions---are not only different to those of good-quality translations but also to most other model-generated translations.


\paragraph{Length-filtering the distributions in $\mathcal{R}_{\text{held}}$.} The results in Table~\ref{tab:ablation_lenfilter} show that length-filtering boosts performance significantly. This is expected: our translation-based length-filtering penalizes translations whose length is anomalous for their respective source sequences. This is particularly useful for detecting oscillatory hallucinations.

\section{Ablations}
\label{app:ablations}
We perform ablations on \textsf{Wass-to-Data} and \textsf{Wass-Combo} for all relevant hyperparameters: the length-filtering parameter $\delta$, the maximum cardinality of $\mathcal{R}$, $|\mathcal{R}|_{\text{max}}$, the value of $k$ to compute the $\textsf{Wass-to-Data}$ scores (step 4 in Figure~\ref{fig:diagramcomputation}), and the threshold on \textsf{Wass-to-Unif} scores to compute \textsf{Wass-Combo} scores. The results are shown in Table~\ref{tab:app_ablation_lenfilter} to Table~\ref{tab:app_ablation_percentileK}, respectively. We also report in Table~\ref{tab:app_costfunction} the performance of \textsf{Wass-to-Data} with a 0/1 cost function instead of the $\ell_1$ distance function.

\paragraph{On length-filtering.} The results in Table~\ref{tab:app_ablation_lenfilter} show that, generally, the bigger the length window, the worse the performance. This is expected: if the test translation is very different in length to those obtained for the source sequences in $\mathcal{R}_{\bm{x}}$, the more penalized it may be for the length mismatch instead of source attention distribution pattern anomalies.

\begin{table}[t]
\centering
\renewcommand\arraystretch{1.0}
\footnotesize
\begin{tabular}{>{\arraybackslash}m{2.45cm} r r}
\toprule
\textsc{Ablation} & AUROC $\uparrow$ & FPR@90TPR $\downarrow$\\
\midrule 
Random Sampling & 80.65$_{\scriptsize\textcolor{gray}{\, 0.15}}$ & 57.06$_{\scriptsize\textcolor{gray}{\, 2.04}}$
\\\cdashlinelr{1-3}\noalign{\vskip 0.5ex} 
\multicolumn{3}{l}{\textbf{\textit{Length Filtering ($\delta > 0$)}}} \\
$\delta = 0.1$ & 84.20$_{\scriptsize\textcolor{gray}{\, 0.15}}$ & 48.15$_{\scriptsize\textcolor{gray}{\, 0.54}}$ \\
$\delta = 0.2$ & 84.37$_{\scriptsize\textcolor{gray}{\, 0.17}}$ & 47.12$_{\scriptsize\textcolor{gray}{\, 1.04}}$ \\
$\delta = 0.3$ & 83.93$_{\scriptsize\textcolor{gray}{\, 0.18}}$ & 48.45$_{\scriptsize\textcolor{gray}{\, 2.32}}$ \\
$\delta = 0.4$ & 83.06$_{\scriptsize\textcolor{gray}{\, 0.16}}$ & 50.12$_{\scriptsize\textcolor{gray}{\, 1.29}}$ \\
$\delta = 0.5$ & 82.78$_{\scriptsize\textcolor{gray}{\, 0.34}}$ & 50.89$_{\scriptsize\textcolor{gray}{\, 0.71}}$\\
\bottomrule
\end{tabular}
\caption{Ablation on \textsf{Wass-to-Data} by changing the length-filtering window to construct $\mathcal{R}$. We present the mean and standard deviation (in subscript) across five random seeds.}
\label{tab:app_ablation_lenfilter}
\end{table}

\paragraph{On the choice of $|\mathcal{R}|_{\text{max}}$.} Table~\ref{tab:app_ablation_Rmax} shows that increasing $|\mathcal{R}|_{\text{max}}$ leads to better performance, with reasonable gains obtained until $|\mathcal{R}|_{\text{max}}=2000$. While this increase in performance may be desirable, it comes at the cost of higher runtime. 

\begin{table}[t]
\centering
\renewcommand\arraystretch{1.0}
\footnotesize
\begin{tabular}{>{\arraybackslash}m{2.45cm} r r}
\toprule
\textsc{Ablation} & AUROC $\uparrow$ & FPR@90TPR $\downarrow$\\
\midrule 
$|\mathcal{R}|_{\text{max}} = 100$ & 82.99$_{\scriptsize\textcolor{gray}{\, 0.19}}$ & 50.86$_{\scriptsize\textcolor{gray}{\, 0.95}}$ \\
$|\mathcal{R}|_{\text{max}} = 500$ & 83.93$_{\scriptsize\textcolor{gray}{\, 0.08}}$ & 48.07$_{\scriptsize\textcolor{gray}{\, 1.37}}$ \\
$|\mathcal{R}|_{\text{max}} = 1000$ & 84.20$_{\scriptsize\textcolor{gray}{\, 0.15}}$ & 48.15$_{\scriptsize\textcolor{gray}{\, 0.54}}$ \\
$|\mathcal{R}|_{\text{max}} = 2000$ & 84.40$_{\scriptsize\textcolor{gray}{\, 0.14}}$ & 49.23$_{\scriptsize\textcolor{gray}{\, 1.08}}$ \\
$|\mathcal{R}|_{\text{max}} = 5000$ & 84.43$_{\scriptsize\textcolor{gray}{\, 0.13}}$ & 48.05$_{\scriptsize\textcolor{gray}{\, 0.59}}$ \\
\bottomrule
\end{tabular}
\caption{Ablation on \textsf{Wass-to-Data} by changing the maximum cardinality of $\mathcal{R}$, $|\mathcal{R}|_{\text{max}}$. We present the mean and standard deviation (in subscript) across five random seeds.}
\label{tab:app_ablation_Rmax}
\end{table}

\paragraph{On the choice of $k$.} The results in Table~\ref{tab:ablation_bottomk} show that the higher the value of $k$, the worse the performance. However, we do not recommend using the minimum distance ($k=1$) as it can be unstable.

\begin{table}[t]
\centering
\renewcommand\arraystretch{1.0}
\footnotesize
\begin{tabular}{>{\arraybackslash}m{2.45cm} r r}
\toprule
\textsc{Ablation} & AUROC $\uparrow$ & FPR@90TPR $\downarrow$\\
\midrule 
Minimum & 84.00$_{\scriptsize\textcolor{gray}{\, 0.33}}$ & 52.03$_{\scriptsize\textcolor{gray}{\, 1.28}}$ \\\cdashlinelr{1-3}\noalign{\vskip 0.5ex}   
\multicolumn{3}{l}{\textbf{\textit{Bottom-k}} ($k > 1$)} \\
$k = 2$ & 84.25$_{\scriptsize\textcolor{gray}{\, 0.23}}$ & 50.07$_{\scriptsize\textcolor{gray}{\, 0.70}}$ \\
$k = 4$ & 84.20$_{\scriptsize\textcolor{gray}{\, 0.15}}$ & 48.15$_{\scriptsize\textcolor{gray}{\, 0.54}}$ \\
$k = 8$ & 83.99$_{\scriptsize\textcolor{gray}{\, 0.08}}$ & 48.38$_{\scriptsize\textcolor{gray}{\, 1.10}}$ \\
$k = 16$ & 83.64$_{\scriptsize\textcolor{gray}{\, 0.04}}$ & 48.05$_{\scriptsize\textcolor{gray}{\, 1.10}}$ \\
$k = 32$ & 83.23$_{\scriptsize\textcolor{gray}{\, 0.07}}$ & 47.34$_{\scriptsize\textcolor{gray}{\, 0.94}}$ \\
\bottomrule
\end{tabular}
\caption{Ablation on \textsf{Wass-to-Data} by obtaining the score $s_{\textsf{wtd}}$ by averaging the bottom-$k$ distances in $\mathcal{R}$ for different values of $k$. We present the mean and standard deviation (in subscript) across five random seeds.}
\label{tab:ablation_bottomk}
\end{table}

\begin{table}[t]
\centering
\renewcommand\arraystretch{1.0}
\footnotesize
\begin{tabular}{>{\arraybackslash}m{2.45cm} r r}
\toprule
\textsc{Ablation} & AUROC $\uparrow$ & FPR@90TPR $\downarrow$\\
\midrule 
$\tau = P_{99}$ & 85.79$_{\scriptsize\textcolor{gray}{\, 0.08}}$ & 51.09$_{\scriptsize\textcolor{gray}{\, 0.97}}$ \\
$\tau = P_{99.5}$ & 86.34$_{\scriptsize\textcolor{gray}{\, 0.07}}$ & 49.64$_{\scriptsize\textcolor{gray}{\, 1.71}}$ \\
$\tau = P_{99.9}$ & 87.17$_{\scriptsize\textcolor{gray}{\, 0.07}}$ & 47.56$_{\scriptsize\textcolor{gray}{\, 1.30}}$ \\
$\tau = P_{99.99}$ & 84.69$_{\scriptsize\textcolor{gray}{\, 0.15}}$ & 48.15$_{\scriptsize\textcolor{gray}{\, 0.54}}$ \\
\bottomrule
\end{tabular}
\caption{Ablation on \textsf{Wass-Combo} by obtaining the score $s_{\textsf{wc}}$ for different scalar thresholds $\tau = P_K$ ($K$-th percentile of $\mathbb{W}_{\textsf{wtu}}$). We present the mean and standard deviation (in subscript) across five random seeds.}
\label{tab:app_ablation_percentileK}
\end{table}

\begin{table}[t]
\centering
\renewcommand\arraystretch{1.0}
\footnotesize
\begin{tabular}{>{\arraybackslash}m{2.45cm} r r}
\toprule
\textsc{Cost Function} & AUROC $\uparrow$ & FPR@90TPR $\downarrow$\\
\midrule 
$\ell_1$ (Wasserstein-1) & 84.20$_{\scriptsize\textcolor{gray}{\, 0.15}}$ & 48.15$_{\scriptsize\textcolor{gray}{\, 0.54}}$ \\
$0/1$ cost & 81.78$_{\scriptsize\textcolor{gray}{\, 0.20}}$ & 51.72$_{\scriptsize\textcolor{gray}{\, 1.17}}$ \\
\bottomrule
\end{tabular}
\caption{Ablation on \textsf{Wass-to-Data} by changing the cost function in the computation of the Wasserstein Distances in Equation~\ref{eq:set_W}.} 
\label{tab:app_costfunction}
\end{table}

\begin{table}[t]
\centering
\renewcommand\arraystretch{1.0}
\footnotesize
\begin{tabular}{>{\arraybackslash}m{2.95cm} r r}
\toprule
\textsc{Ablation} & AUROC $\uparrow$ & FPR@90TPR $\downarrow$\\
\midrule 
Our \textsf{Wass-Combo} & 87.17$_{\scriptsize\textcolor{gray}{\, 0.07}}$ & 47.56$_{\scriptsize\textcolor{gray}{\,1.30}}$\\\cdashlinelr{1-3}\noalign{\vskip 0.5ex} 
\textsf{Wass-to-Unif} ($\lambda=0$) & 80.37 & 72.22\\
$\lambda = 0.2$ & 81.57$_{\scriptsize\textcolor{gray}{\, 0.00}}$ & 69.02$_{\scriptsize\textcolor{gray}{\,0.19}}$ \\
$\lambda = 0.4$ & 82.28$_{\scriptsize\textcolor{gray}{\, 0.01}}$ & 68.69$_{\scriptsize\textcolor{gray}{\, 0.13}}$ \\
$\lambda = 0.6$ & 82.77$_{\scriptsize\textcolor{gray}{\, 0.02}}$ & 66.15$_{\scriptsize\textcolor{gray}{\, 1.09}}$ \\
$\lambda = 0.8$ & 83.48$_{\scriptsize\textcolor{gray}{\, 0.05}}$ & 63.01$_{\scriptsize\textcolor{gray}{\, 0.44}}$ \\
\textsf{Wass-to-Data} ($\lambda=1$) & 84.20$_{\scriptsize\textcolor{gray}{\, 0.15}}$ & 48.15$_{\scriptsize\textcolor{gray}{\, 0.54}}$ \\
\bottomrule
\end{tabular}
\caption{Convex combination of \textsf{Wass-to-Unif} and \textsf{Wass-to-Data} scores. We present the mean and standard deviation (in subscript) across five random seeds.}
\label{tab:ablation_wasscombo_convex}
\end{table}

\begin{table}[t]
\centering
\renewcommand\arraystretch{1.0}
\footnotesize
\begin{tabular}{>{\arraybackslash}m{2.45cm} r r}
\toprule
\textsc{Method} & AUROC $\uparrow$ & FPR@90TPR $\downarrow$\\\midrule 
\multicolumn{3}{l}{\textbf{\textit{All}}} \\
\textsf{ALTI+} & 84.27 & 66.30 \\
\textsf{Wass-Combo} & \textbf{87.17}$_{\scriptsize\textcolor{gray}{\, 0.07}}$ & \textbf{47.56}$_{\scriptsize\textcolor{gray}{\, 1.30}}$ \\\cdashlinelr{1-3}\noalign{\vskip 0.5ex}   
\multicolumn{3}{l}{\textbf{\textit{Fully detached}}} \\
\textsf{ALTI+} & \textbf{98.21} & \textbf{2.15} \\
\textsf{Wass-Combo} & 96.57$_{\scriptsize\textcolor{gray}{\, 0.10}}$ & 3.56$_{\scriptsize\textcolor{gray}{\, 0.00}}$ \\\cdashlinelr{1-3}\noalign{\vskip 0.5ex}  
\multicolumn{3}{l}{\textbf{\textit{Oscillatory}}} \\
\textsf{ALTI+} & 71.39 & 76.72 \\
\textsf{Wass-Combo} & \textbf{85.74}$_{\scriptsize\textcolor{gray}{\, 0.10}}$ & \textbf{41.38}$_{\scriptsize\textcolor{gray}{\, 1.59}}$ \\\cdashlinelr{1-3}\noalign{\vskip 0.5ex}  
\multicolumn{3}{l}{\textbf{\textit{Strongly Detached}}} \\ 
\textsf{ALTI+} & 73.77 & 89.41 \\
\textsf{Wass-Combo} & \textbf{78.89}$_{\scriptsize\textcolor{gray}{\, 0.15}}$ & \textbf{64.55}$_{\scriptsize\textcolor{gray}{\, 1.93}}$ \\
\bottomrule
\end{tabular}
\caption{Comparison between \textsf{ALTI+} and \textsf{Wass-Combo} detection methods. We present the mean and standard
deviation (in subscript) across five random seeds.} 
\label{tab:alti}
\end{table}

\paragraph{On the choice of threshold on \textsf{Wass-to-Unif} scores.} Table~\ref{tab:app_ablation_percentileK} show that, generally, a higher threshold $\tau$ leads to a better performance of \textsf{Wass-Combo}. \textsf{Wass-to-Unif} scores are generally very high for fully detached hallucinations, a type of hallucinations that \textsf{Wass-to-Data} struggles more to detect. Thus, when combined in \textsf{Wass-Combo}, we obtain significant boosts in overall performance. However, if the threshold on \textsf{Wass-to-Unif} scores is set too low, \textsf{Wass-to-Combo} will correspond to \textsf{Wass-to-Unif} more frequently which may not be desirable as \textsf{Wass-to-Data} outperforms it for all other types of hallucinations. If set too high, fewer fully detached hallucinations may pass that threshold and may then be misidentified with \textsf{Wass-to-Data} scores. 

\paragraph{On the choice of \textsf{Wass-to-Data} cost function.} Table~\ref{tab:app_costfunction} shows that using the $\ell_1$ cost function instead of using the 0/1 cost function to compute \textsf{Wass-to-Data} scores leads to significant improvements. This suggests that when comparing the source mass attention distribution of a test translation to other such distributions obtained for other translations (instead of the ad-hoc uniform distribution used for \textsf{Wass-to-Unif} scores), the information from the location of the source attention mass is helpful to obtain better scores.

\paragraph{On the formulation of \textsf{Wass-Combo}.} To combine the information from \textsf{Wass-to-Unif} and \textsf{Wass-to-Data}, we could also perform a convex combination of the two scores:
\begin{align}\label{eq:wass_combo_alternative}
    s_{\textsf{wc}}(\bm{x}) = \lambda s_{\textsf{wtd}}(\bm{x}) + (1-\lambda) \tilde{s}_{\textsf{wtu}}(\bm{x})
\end{align}
for a predefined scalar parameter $\lambda$. In Table~\ref{tab:ablation_wasscombo_convex}, we show that this method is consistently subpar to our two-pass approach. In fact, this linear interpolation is not able to bring additional gains in performance for any of the tested parameters $\lambda$ when compared to \textsf{Wass-to-Data}.

\begin{figure*}[t]
    \centering
    \includegraphics[width=\linewidth]{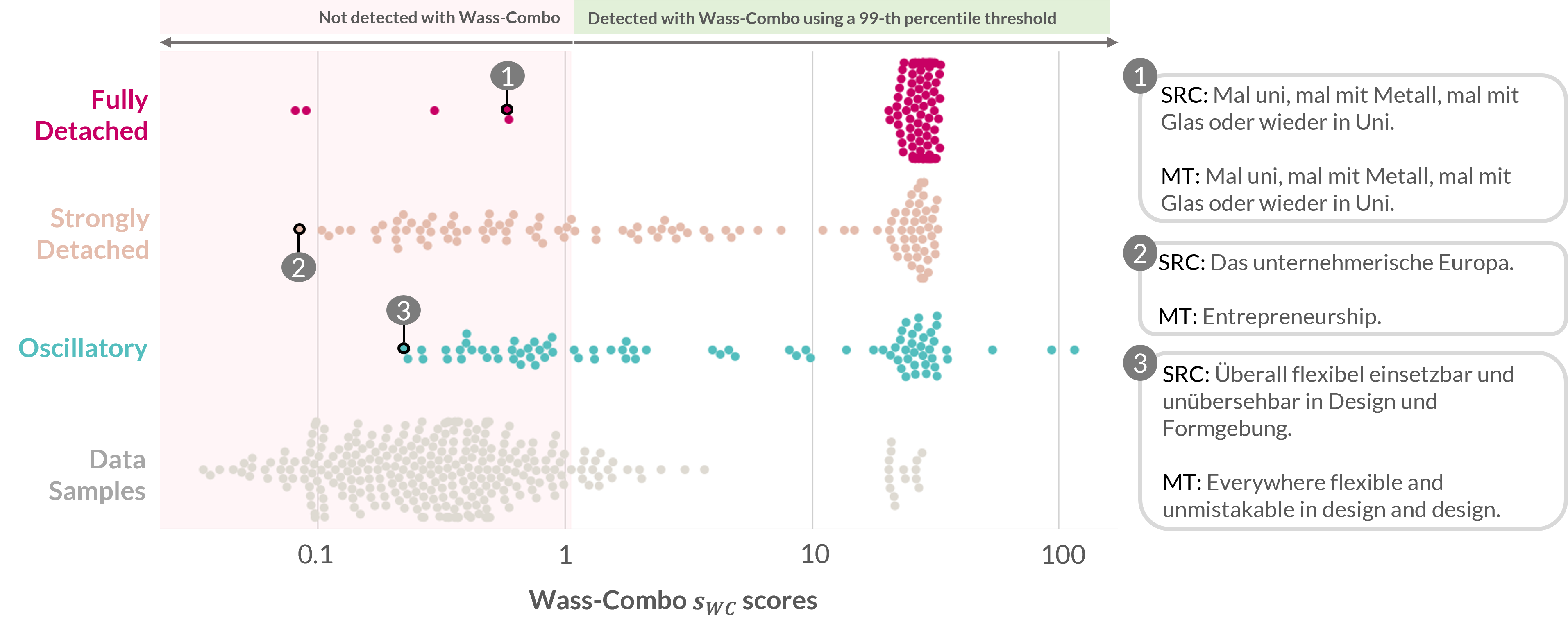}
    \caption{Distribution of \textsf{Wass-Combo} scores (on log-scale) for each type of hallucination and performance on a fixed-threshold scenario. We highlight three hallucinations that are not detected by \textsf{Wass-Combo}. These represent hallucinations in the dataset that it struggles to identify: (1) exact copies of the source sequence, (2) small level of detachment in strongly detached hallucinations, and (3) mild repetitions of 1-grams ("design").}
    \label{fig:errors_halls_appendix}
\end{figure*}

\begin{table*}[h!]
    \footnotesize
    \centering
    \renewcommand\arraystretch{1.05}
    \begin{tabular}{>{\arraybackslash}m{0.15\textwidth} >{\arraybackslash}m{0.8\textwidth}}
    \toprule
        \multicolumn{2}{c}{\textsc{\textbf{Oscillatory Hallucinations not detected with Wass-Combo}}} \\ \midrule
        \textsc{Source} &  Überall flexibel einsetzbar und unübersehbar in Design und Formgebung.\\
        \textsc{Translation} & Everywhere flexible and unmistakable in \textcolor{HallRedText}{\hl{design}} and \textcolor{HallRedText}{\hl{design}}.\\ \arrayrulecolor{gray!30}\midrule

        \textsc{Source} &  Um kahlen Stellen, wenn sie ohne Rüstung pg.\\
        \textsc{Translation} & To \textcolor{HallRedText}{\hl{dig dig}} digits if they have no armor pg.\\ \arrayrulecolor{gray!30}\midrule

        \textsc{Source} &  Damit wird, wie die Wirtschaftswissenschaftler sagen, der Nennwert vorgezogen.\\
        \textsc{Translation} & This, as economists say, puts \textcolor{HallRedText}{\hl{the par value}} before \textcolor{HallRedText}{\hl{the par value}}.\\ \arrayrulecolor{gray!30}\midrule

        \textsc{Source} &  Besonders beim Reinigen des Verflüssigers kommt Ihnen dies zugute.\\
        \textsc{Translation} & Especially when cleaning the \textcolor{HallRedText}{\hl{liquefied liquefied liquefied}}.\\ \arrayrulecolor{gray!30}\midrule

        \textsc{Source} &  Müssen die Verkehrsmittel aus- oder abgewählt werden ?\\
        \textsc{Translation} & Do you need to \textcolor{HallRedText}{\hl{opt-out}} or \textcolor{HallRedText}{\hl{opt-out}} of transport?\\ \arrayrulecolor{gray!30}\midrule

        \textsc{Source} &  Schnell drüberlesen - "Ja" auswählen und weiter gehts.\\
        \textsc{Translation} & Simply \textcolor{HallRedText}{\hl{press the "Yes"}} button and \textcolor{HallRedText}{\hl{press the "Yes."}}\\ \arrayrulecolor{gray!30}\midrule

        \textsc{Source} &  Auf den jeweiligen Dorfplätzen finden sich Alt und Jung zum Schwätzchen und zum Feiern zusammen.\\
        \textsc{Translation} & Old and young people will find themselves together in the village's respective squares for \textcolor{HallRedText}{\hl{fun and fun}}.\\ \arrayrulecolor{gray!30}\midrule

        \textsc{Source} &  Zur Absicherung der E-Mail-Kommunikation auf Basis von PGP- als auch X.509-Schlüsseln hat die Schaeffler Gruppe eine Zertifizierungsinfrastruktur (Public Key Infrastructure PKI) aufgebaut.\\
        \textsc{Translation} & The Schaeffler Group has set up a \textcolor{HallRedText}{\hl{Public Key Infrastructure PKI (Public Key Infrastructure PKI)}} to secure e-mail communication based on PGP and X.509 keys.\\ \bottomrule
        
    \end{tabular}\vspace{5pt}
    
    \caption{Examples of oscillatory hallucinations randomly sampled from the set of oscillatory hallucinations not detected with \textsf{Wass-Combo}. Most hallucinations come in the form of mild repetitions of 1-grams or 2-grams.}
    \label{tab:osc_not_detected_w_wc}
\end{table*}

\section{Analysis against ALTI+}
\label{app:alti}
Concurrently to our work, \citet{dale2022detecting} leveraged \textsf{ALTI+}~\citep{altijavier2022}, a method that evaluates the global relative contributions of
both source and target prefixes to model predictions, for detection of hallucinations. As hallucinations are translations detached from the source sequence, \textsf{ALTI+} is able to detect them by identifying sentences with minimal source contribution. In Table~\ref{tab:alti}, we show that \textsf{ALTI+} slightly outperforms \textsf{Wass-Combo} for fully detached hallucinations, but lags considerably behind on what comes to detecting strongly detached and oscillatory hallucinations. 

\section{Error Analysis of \textsf{Wass-Combo}}\label{app:error_analysis}
We show a qualitative analysis on the same fixed-threshold scenario described in Section~\ref{sec:type_analysis} in Figure~\ref{fig:errors_halls_appendix}. Differently to Figure~\ref{fig:errors_halls}, we provide examples of translations that have not been detected by \textsf{Wass-Combo} for the chosen threshold. 

Our detector is not able to detect fully detached hallucinations that come in the form of exact copies of the source sentence. For these pathological translations, the attention map is mostly diagonal and is thus not anomalous. Although these are severe errors, we argue that, in a real-world application, such translations can be easily detected with string matching heuristics. 

We also find that our detector \textsf{Wass-Combo} struggles with oscillatory hallucinations that come in the form of mild repetitions of 1-grams or 2-grams (see example in Figure~\ref{fig:errors_halls_appendix}). To test this hypothesis, we implemented the binary heuristic \textsf{top n-gram count}~\citep{raunak-etal-2021-curious, guerreiroetal2022} to verify whether a translation is a severe oscillation: given the entire $\mathcal{D}_{\text{held}}$, a translation is flagged as an oscillatory hallucination if (i) it is in the set of 1\% lowest-quality translations according to \textsf{CometKiwi} and (ii) the count of the top repeated 4-gram in the translation is greater than the count of the top repeated source 4-gram by at least 2. Indeed, more than $90\%$ of the oscillatory hallucinations not detected by \textsf{Wass-Combo} in Figure~\ref{fig:errors_halls_appendix} were not flagged by this heuristic. We provide 8 examples randomly sampled from the set of oscillatory hallucinations not detected with \textsf{Wass-Combo} in Table~\ref{tab:osc_not_detected_w_wc}. Close manual evaluation of these hallucinations further backs the hypothesis above.

\begin{table*}[h!]
\centering
\renewcommand\arraystretch{1}
\footnotesize
\begin{tabular}{>{\arraybackslash}m{2.45cm} r r >{\arraybackslash}m{0.02cm} r r}
\toprule
\multirow{2}{*}{\textsc{Detector}} & \multicolumn{2}{c}{\textbf{\textsc{ro-en}}} & & \multicolumn{2}{c}{\textbf{\textsc{ne-en}}} \\\cmidrule{2-3}\cmidrule{5-6}
& AUROC $\uparrow$ & FPR@90TPR $\downarrow$ & & AUROC $\uparrow$ & FPR@90TPR $\downarrow$ \\
\midrule 
\multicolumn{6}{l}{\textbf{\textit{External Detectors}}} \\
{\textsf{CometKiwi}} $^{\dagger}$ & {99.62} & \textbf{0.49} & & \textbf{97.64} & \textbf{4.03} 
\\
\textsf{LaBSE} & \textbf{99.72} & \textbf{0.49} & & 92.34  & 20.03 \\\cdashlinelr{1-6}\noalign{\vskip 0.5ex}   
\multicolumn{6}{l}{\textbf{\textit{Model-based Detectors}}} \\
\textsf{Attn-ign-SRC}   & 99.16 & 0.93 & & 28.66 & 100.0 \\
\textsf{Seq-Logprob}    & 91.97 & 16.42 & & 26.38 & 99.94\\[0.075cm] 
\rowcolor{gray!10} \multicolumn{6}{l}{\textbf{\textsc{Ours}}}\\\noalign{\vskip 0.25ex} 
\textsf{Wass-to-Unif} & \textbf{99.30} & \textbf{0.46} & & 81.49 & 64.23 \\
\textsf{Wass-to-Data} & 96.54$_{\scriptsize\textcolor{gray}{\, 0.07}}$ & {10.36}$_{\scriptsize\textcolor{gray}{\, 0.30}}$ & & \textbf{90.18}$_{\scriptsize\textcolor{gray}{\, 0.13}}$ & \textbf{48.52}$_{\scriptsize\textcolor{gray}{\, 2.64}}$\\
\textsf{Wass-Combo} & 98.75$_{\scriptsize\textcolor{gray}{\, 0.06}}$ & \textbf{0.46}$_{\scriptsize\textcolor{gray}{\, 0.00}}$ & & \textbf{90.16}$_{\scriptsize\textcolor{gray}{\, 0.13}}$ & \textbf{48.52}$_{\scriptsize\textcolor{gray}{\, 2.64}}$\\
\bottomrule
\end{tabular}
\caption{Performance of all hallucination detectors. For \textsf{Wass-to-Data} and \textsf{Wass-Combo} We present the mean and standard deviation (in subscript) scores across five random seeds. $^{\dagger}$\textsf{CometKiwi} has been trained on these MLQE-PE data samples.}
\label{tab_app:auroc_fpr_new_lps}
\end{table*}

\section{Experiments on the MLQE-PE dataset}
\label{app:mlqepe_experiments}
In order to establish the broader validity of our model-based detectors, we present an analysis on their performance for other NMT models and on mid and low-resource language pairs. Overall, the detectors exhibit similar trends to those discussed in the main text~(Section~\ref{sec:results}).

\subsection{Model and Data}
\label{app:modelanddata}
The dataset from~\cite{guerreiroetal2022} analysed in the main text is the only available dataset that contains human annotations of hallucinated translations. Thus, in this analysis we will have to make use of other human annotations to infer annotations for hallucinations. For that end, we follow a similar setup to that of~\cite{tangspeciahallreduction2022} and use the MLQE-PE dataset~\citep{fomicheva-etal-2022-mlqe}--- that has been reported to contain low-quality translations and hallucinations for \textsc{ne-en} and \textsc{ro-en}~\citep{specia-etal-2021-findings}--- to test the performance of our detectors on these language pairs. 

The \textsc{ne-en} and \textsc{ro-en} MLQE-PE datasets contain 7000 translations and their respective human quality assessments (from 1 to 100). Each translation is scored by three different annotators. As hallucinations lie at the extreme end of NMT pathologies~\citep{raunak-etal-2021-curious}, we consider a translation to be a hallucination if at least two annotators (majority) gave it a quality score of 1.\footnote{We tried other methods to infer hallucinations from the annotations (e.g. average quality score below 5, at least one quality score of 1). The trends on performance were similar to those reported in this section.} This process leads to 30 hallucinations for \textsc{ne-en} and 237 hallucinations for \textsc{ro-en}. Although the number of hallucinations for \textsc{ne-en} is relatively small, we decide to also report experiments on this language pair because the type of hallucinations found for \textsc{ne-en} is very different to those found for \textsc{ro-en}: almost all \textsc{ne-en} hallucinations are oscillatory, whereas almost all \textsc{ro-en} are fully detached. 

To obtain all model-based information required to build the detectors, we use the same Transformer models that generated the translations in the datasets in consideration. All details can be found in~\citet{fomicheva-etal-2022-mlqe} and the official project repository\footnote{\url{https://github.com/facebookresearch/mlqe}}. Moreover, to build our held-out databases of source mass distributions, we used readily available Europarl data~\citep{koehn-2005-europarl} for \textsc{ro-en} ($\sim$100k samples), and filtered Nepali Wikipedia monolingual data\footnote{Note that creating the datastore only requires access to monolingual data. If quality filtering is needed and references are not available, we suggest using quality estimation or cross-lingual embedding similarity models to filter low-quality translations.} used in~\citep{koehn-etal-2019-findings} for \textsc{ne-en} ($\sim$80k samples).

\subsection{Results}
\paragraph{The trends in Section~\ref{sec:onthefly_analysis} hold for other language pairs.} The results in Table~\ref{tab_app:auroc_fpr_new_lps} establish the broader validity of our detectors for other NMT models and, importantly, for mid and low-resource language pairs. Similarly to the analysis in~\ref{sec:onthefly_analysis}, we find that our detectors (i) exhibit better performance than other model-based detectors with significant gains on the low-resource \textsc{ne-en} language pair; and (ii) can be competitive with external detectors that leverage large models.

\paragraph{The trends in Section~\ref{sec:type_analysis} hold for other language pairs.} In Section~\ref{app:model_data_details}, we remark that almost all \textsc{ne-en} hallucinations are oscillatory, whereas almost all \textsc{ro-en} hallucinations are fully detached. With that in mind, the results in Table~\ref{tab_app:auroc_fpr_new_lps} establish the validity of the claims in the main-text (Section~\ref{sec:type_analysis}) on these language pairs: (i) detecting fully detached hallucinations is remarkably easy for most detectors, and \textsf{Wass-to-Unif} outperforms \textsf{Wass-to-Data} on this type of hallucinations (see results for \textsc{ro-en}); and (ii) our detectors significantly outperform all previous model-based detectors on detecting oscillatory hallucinations (see results for \textsc{ne-en}), which further confirms the notion that some detectors specialize on different types of hallucinations (e.g \textsf{Attn-ign-SRC} is particularly fit for detecting fully detached hallucinations, but it does not work for oscillatory hallucinations).